\documentclass[letterpaper]{article}
\usepackage{aaai19}
\usepackage{times}
\usepackage{helvet}
\usepackage{courier}
\usepackage{url}
\usepackage{graphicx}
\usepackage{array}
\usepackage{subcaption}
\usepackage[dvipsnames]{xcolor}
\usepackage{tikz}
\usetikzlibrary{calc}
\usepackage{booktabs}       %
\usepackage{dsfont}
\usepackage{graphicx}
\usepackage{subcaption}
\usepackage{textcomp}  %

\definecolor{ramiblue}{HTML}{007bd5}
\definecolor{ramigray}{HTML}{a9a9a9}
\definecolor{ramired}{HTML}{c11b24}

\newcommand{\texteight}{\texttt{text8}}
\newcommand{\enwikeight}{\texttt{enwik8}}
\newcommand{\enwiknine}{\texttt{enwik9}}
\newcommand{\lmoneb}{\texttt{lm1b}}
\newcommand{\bigmodel}{\textsc{T64}}
\newcommand{\smallmodel}{\textsc{T12}}
\newcommand{\todo}[1]{{\footnotesize\color{red}{\textbf{#1}}}}
\newcommand{\comment}[1]{}
\newcommand{\corpus}[1]{\begin{quote}{\small \emph{#1}}\end{quote}}
\newcommand{\newcite}[1]{\citeauthor{#1}~(\citeyear{#1})}
\newcommand*\svtick{\textquotesingle}
\newcommand*\dvtick{\textquotesingle\textquotesingle}
\newcommand*\tvtick{\textquotesingle\textquotesingle\textquotesingle}

\title{Character-Level Language Modeling with Deeper Self-Attention}

\author{Rami Al-Rfou\thanks{Equal contribution.}\\
 \And
 Dokook Choe\footnotemark[1]
 \And 
 Noah Constant\footnotemark[1]
 \\\hspace{0.5pt}
 \\
 Google \textsc{AI} Language
 \\
 \texttt{\{rmyeid, choed, nconstant, xyguo, llion\}@google.com}
 \And
 Mandy Guo\footnotemark[1]
 \And
 Llion Jones\footnotemark[1]
}

\begin{document}

\maketitle

\begin{abstract}

LSTMs and other RNN variants have shown strong performance on character-level language modeling.
These models are typically trained using truncated backpropagation through time, and it is common to assume that their success stems from their ability to remember long-term contexts.
In this paper, we show that a deep (64-layer) transformer model \cite{NIPS2017_7181} with fixed context outperforms RNN variants by a large margin, achieving state of the art on two popular benchmarks: 1.13 bits per character on \texteight{} and 1.06 on \enwikeight{}.
To get good results at this depth, we show that it is important to add auxiliary losses, both at intermediate network layers and intermediate sequence positions.

\end{abstract}

\section{Introduction}

Character-level modeling of natural language text is challenging, for several reasons.
First, the model must learn a large vocabulary of words ``from scratch''.
Second, natural text exhibits dependencies over long distances of hundreds or thousands of time steps.
Third, character sequences are longer than word sequences and thus require significantly more steps of computation.

In recent years, strong character-level language models typically follow a common template \cite{conf/interspeech/MikolovKBCK10,extensionsmikolov,sundermeyer2012lstm}.
A recurrent neural net (RNN) is trained over mini-batches of text sequences, using a relatively short sequence length (e.g.~200 tokens).
To capture context longer than the batch sequence length, training batches are provided in sequential order, and the hidden states from the previous batch are passed forward to the current batch.
This procedure is known as ``truncated backpropagation through time'' (TBTT), because the gradient computation doesn't proceed further than a single batch \cite{werbos1990backpropagation}.
A range of methods have arisen for unbiasing and improving TBTT \cite{tallec2017unbiasing,ke2017sparse}.

While this technique gets good results, it adds complexity to the training procedure, and recent work suggests that models trained in this manner don't actually make ``strong'' use of long-term context.
For example \newcite{khandelwal2018lm} find that a word-based LSTM language model only effectively uses around 200 tokens of context (even if more is provided), and that word order only has an effect within approximately the last 50 tokens.

In this paper, we show that a non-recurrent model can achieve strong results on character-level language modeling.
Specifically, we use a deep network of transformer self-attention layers \cite{NIPS2017_7181} with causal (backward-looking) attention to process fixed-length inputs and predict upcoming characters.
The model is trained on mini-batches of sequences from random positions in the training corpus, with no information passed from one batch to the next.

Our primary finding is that the transformer architecture is well-suited to language modeling over long sequences and could replace RNNs in this domain.
We speculate that the transformer's success here is due to its ability to ``quickly'' propagate information over arbitrary distances; by comparison, RNNs need to learn to pass relevant information forward step by step.

We also find that some modifications to the basic transformer architecture are beneficial in this domain.
Most importantly, we add three auxiliary losses, requiring the model to predict upcoming characters (i) at intermediate sequence positions, (ii) from intermediate hidden representations, and (iii) at target positions multiple steps in the future.
These losses speed up convergence, and make it possible to train deeper networks.

\section{Character Transformer Model}

Language models assign a probability distribution over token sequences $t_{0:L}$ by factoring out the joint probability as follows, where $L$ is the sequence length:

\begin{equation}
    \Pr(t_{0:L}) = P(t_0) \prod_{i=1}^{L} \Pr(t_i|t_{0:i-1}),
\end{equation}

To model the conditional probability $\Pr(t_i|t_{0:i-1})$, we train a transformer network to process the character sequence $t_{0:i-1}$.
Transformer networks have recently showed significant gains in tasks that require processing sequences accurately and efficiently.

Our character-level transformer architecture has 64 transformer layers.
Following \newcite{NIPS2017_7181}, by ``transformer layer'' we mean a block containing a multihead self-attention sub-layer followed by a feed-forward network of two fully connected sub-layers.
For more details on the transformer architecture, refer to \newcite{NIPS2017_7181} and the \texttt{tensor2tensor} library\footnote{\texttt{https://github.com/tensorflow/tensor2tensor}}.
To ensure that the model's predictions are only conditioned on past characters, we mask our attention layers with a causal attention, so each position can only attend leftward.
This is the same as the ``masked attention'' in the decoder component of the original transformer architecture used for sequence-to-sequence problems \cite{NIPS2017_7181}.

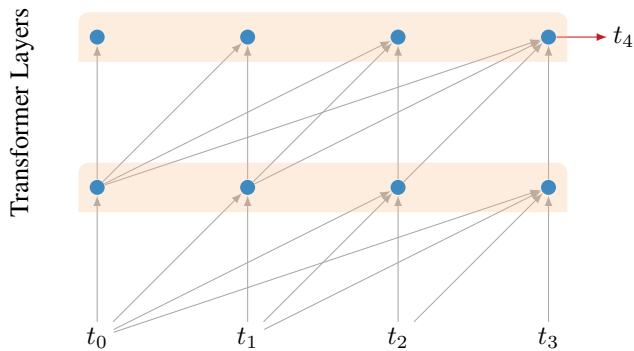
\begin{figure}[htb]
  \centering
  \begin{tikzpicture}
    \coordinate (Origin)   at (0,0);

    \foreach \x in {0, 1,...,3}{
        \node(hidden\x0)[inner sep=2pt] at (2*\x, 0) {$t_{\x}$};

    }

    \foreach \x in {0, 1,...,3}{
      \foreach \y in {1,...,2}{
        \node(hidden\x\y)[circle,inner sep=2pt,fill=ramiblue] at (2*\x,2*\y) {};
      }
    }

    \foreach \x in {0, 1,...,3}{
      \foreach \y [evaluate={\yprev=int(\y-1)}] in {1,..., 2}{
      \foreach \z in {0, ..., \x}{
        \draw[ramigray, -latex] (hidden\z\yprev) -- (hidden\x\y);
        }
      }
    }

    \foreach \x [evaluate={\xprev=int(\x-1)}]  in {4,...,4}{
        \node(output\x)[inner sep=2pt] at (2*\x-1, 4.0) {$t_{\x}$};
    }

    \foreach \x [evaluate={\xprev=int(\x-1)}]  in {4,...,4}{
        \draw[ramired, -latex] (hidden\xprev2) -- (output\x);

    }

  \foreach \y in {1, ..., 2}{
    \draw[name=layer\y, draw=none, fill=Apricot, opacity=0.25, rounded corners]  (-0.25, 2*\y-0.33) -- (-0.25, 2*\y+0.33) --  (6.25, 2*\y+0.33) -- (6.25, 2*\y-0.33);
    }
    
  \node(transformerlabel)[draw=none, align=left, rotate=90] at (-1,3) {Transformer Layers};

  \end{tikzpicture}
  \caption{Character transformer network of two layers processing a four character sequence to predict $t_4$. The causal attention mask limits information to left-to-right flow. Red arrows highlight the prediction task the network has to learn.}
  \label{figure:transformer}
\end{figure}

Figure \ref{figure:transformer} shows our initial model with the causal attention mask limiting information flow from left to right.
Each character prediction is conditioned only on the characters that appeared earlier.

\subsection{Auxiliary Losses}
\label{section:aux}
Our network is, to our knowledge, deeper than any transformer network discussed in previous work.
In initial experiments, we found training a network deeper than ten layers to be challenging, with slow convergence and poor accuracy.
We were able to deepen the network to better effect through the addition auxiliary losses, which sped up convergence of the training significantly.

We add several types of auxiliary losses, corresponding to intermediate positions, intermediate layers, and non-adjacent targets.
We hypothesize that these losses not only speed up convergence but also serve as an additional regularizer.
During training, the auxiliary losses get added to the total loss of the network with discounted weights.
Each type of auxiliary loss has its own schedule of decay.
During evaluation and inference time, only the prediction of the final position at the final layer is used.

One consequence of this approach is that a number of the network parameters are only used during training\textemdash specifically, the parameters in the output classification layers associated with predictions made from intermediate layers and predictions over non-adjacent targets.
Thus, when listing the number of parameters in our models, we distinguish between ``training parameters'' and ``inference parameters''.

\subsubsection{Multiple Positions}

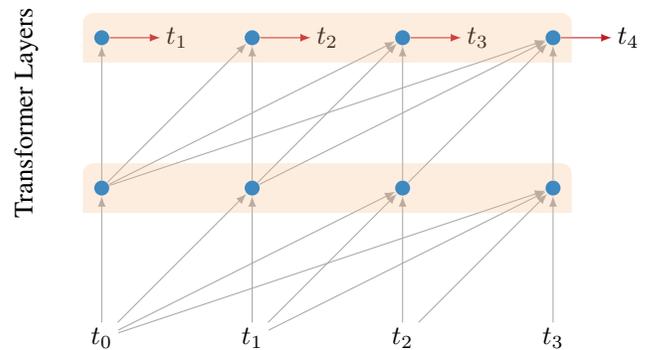
\begin{figure}[htb]
  \centering
  \begin{tikzpicture}
    \coordinate (Origin)   at (0,0);

    \foreach \x in {0, 1,...,3}{
        \node(hidden\x0)[inner sep=2pt] at (2*\x, 0) {$t_{\x}$};

    }

    \foreach \x in {0, 1,...,3}{
      \foreach \y in {1,...,2}{
        \node(hidden\x\y)[circle,inner sep=2pt,fill=ramiblue] at (2*\x,2*\y) {};
      }
    }

    \foreach \x in {0, 1,...,3}{
      \foreach \y [evaluate={\yprev=int(\y-1)}] in {1,..., 2}{
      \foreach \z in {0, ..., \x}{
        \draw[ramigray, -latex] (hidden\z\yprev) -- (hidden\x\y);
        }
      }
    }

    \foreach \x [evaluate={\xprev=int(\x-1)}]  in {1,...,4}{
        \node(output\x)[inner sep=2pt] at (2*\x-1, 4.0) {$t_{\x}$};
    }

    \foreach \x [evaluate={\xprev=int(\x-1)}]  in {1,...,4}{
        \draw[ramired, -latex] (hidden\xprev2) -- (output\x);

    }

  \foreach \y in {1, ..., 2}{
    \draw[name=layer\y, draw=none, fill=Apricot, opacity=0.25, rounded corners]  (-0.25, 2*\y-0.33) -- (-0.25, 2*\y+0.33) --  (6.25, 2*\y+0.33) -- (6.25, 2*\y-0.33);
    }
    
  \node(transformerlabel)[draw=none, align=left, rotate=90] at (-1,3) {Transformer Layers};

  \end{tikzpicture}
  \caption{Adding the intermediate positions prediction tasks to our network. Now, we predict the final character $t_4$ and all intermediate characters $t_{0:3}$. $t_3$ has access only to $t_{0:2}$ because of the causal attention masks. All of these losses contribute equally during training.}
  \label{figure:multiple_positions}
\end{figure}

First, we add prediction tasks for each position in the final layer, extending our predictions from one per example to $|L|$ (sequence length).
Note, predicting over all sequence positions is standard practice in RNN-based approaches.
However in our case, since no information is passed forward across batches, this is forcing the model to predict given smaller contexts---sometimes just one or two characters.
It is not obvious whether these secondary training tasks should help on the primary task of predicting with full context.
However, we find that adding this auxiliary loss speeds up training and gives better results (see Ablation Experiments below).
Figure \ref{figure:multiple_positions} illustrates the task of predicting across all sequence positions.
We add these losses during training without decaying their weights.

\subsubsection{Intermediate Layer Losses}
\label{section:intermediate_layer_losses}

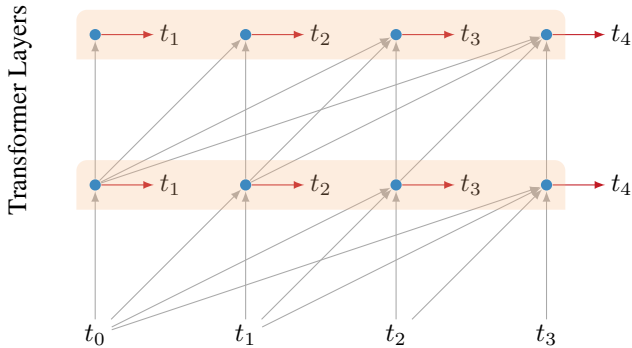
\begin{figure}[hbt]
  \centering
  \begin{tikzpicture}
    \coordinate (Origin)   at (0,0);

    \foreach \x in {0, 1,...,3}{
        \node(hidden\x0)[inner sep=2pt] at (2*\x, 0) {$t_{\x}$};

    }

    \foreach \x in {0, 1,...,3}{
      \foreach \y in {1,...,2}{
        \node(hidden\x\y)[circle,inner sep=1.5pt,fill=ramiblue] at (2*\x,2*\y) {};
      }
    }

    \foreach \x in {0, 1,...,3}{
      \foreach \y [evaluate={\yprev=int(\y-1)}] in {1,..., 2}{
      \foreach \z in {0, ..., \x}{
        \draw[ramigray, -latex] (hidden\z\yprev) -- (hidden\x\y);
        }
      }
    }

    \foreach \x [evaluate={\xprev=int(\x-1)}]  in {1,...,4}{
        \node(output2\x)[inner sep=2pt] at (2*\x-1, 4.0) {$t_{\x}$};
        \draw[ramired, -latex] (hidden\xprev2) -- (output2\x);
    }

    \foreach \x [evaluate={\xprev=int(\x-1)}]  in {1,...,4}{
        \node(output1\x)[inner sep=2pt] at (2*\x-1, 2.0) {$t_{\x}$};
        \draw[ramired, -latex] (hidden\xprev1) -- (output1\x);
    }

  \foreach \y in {1, ..., 2}{
    \draw[name=layer\y, draw=none, fill=Apricot, opacity=0.25, rounded corners]  (-0.25, 2*\y-0.33) -- (-0.25, 2*\y+0.33) --  (6.25, 2*\y+0.33) -- (6.25, 2*\y-0.33);
    }
    
  \node(transformerlabel)[draw=none, align=left, rotate=90] at (-1,3) {Transformer Layers};

  \end{tikzpicture}
  \caption{Our network after adding prediction tasks for the intermediate layers. For this example of two layers, the losses of the intermediate layer prediction tasks will be absent after finishing 25\% of the training.}
  \label{figure:layer_losses}
\end{figure}

In addition to the final prediction layer, we add predictions made from the output of each intermediate transformer layer.
As with the final layer, we add predictions for all intermediate positions in the sequence (see Figure \ref{figure:layer_losses}).
Lower layers are weighted to contribute less and less to the loss as training progresses.
If there are $n$ layers total, then the $l^{th}$ intermediate layer stops contributing any loss after finishing $l/2n$ of the training. 
This schedule drops all intermediate losses after half of the training is done.

\subsubsection{Multiple Targets}

At each position in the sequence, the model makes two (or more) predictions of future characters.
For each new target we introduce a separate classifier.
The losses of the extra targets get weighted by a multiplier of 0.5 before being added to their corresponding layer loss.
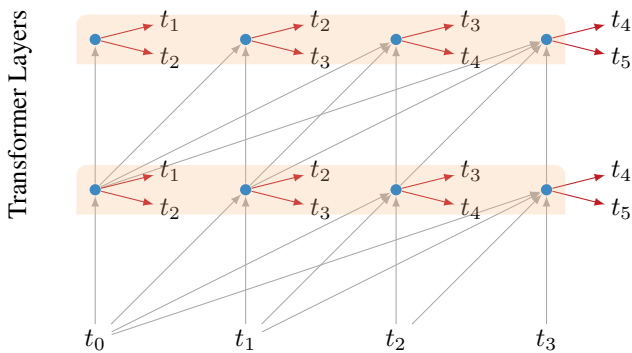
\begin{figure}[htb]
  \centering
  \begin{tikzpicture}
    \coordinate (Origin)   at (0,0);

    \foreach \x in {0, 1,...,3}{
        \node(hidden\x0)[inner sep=2pt] at (2*\x, 0) {$t_{\x}$};

    }

    \foreach \x in {0, 1,...,3}{
      \foreach \y in {1,...,2}{
        \node(hidden\x\y)[circle,inner sep=1.5pt,fill=ramiblue] at (2*\x,2*\y) {};
      }
    }

    \foreach \x in {0, 1,...,3}{
      \foreach \y [evaluate={\yprev=int(\y-1)}] in {1,..., 2}{
      \foreach \z in {0, ..., \x}{
        \draw[ramigray, -latex] (hidden\z\yprev) -- (hidden\x\y);
        }
      }
    }

    \foreach \x [evaluate={\xprev=int(\x-1)}, evaluate={\xnext=int(\x+1)}]  in {1,...,4}{
        \node(output2\x)[inner sep=2pt] at (2*\x-1, 4.25) {$t_{\x}$};
        \draw[ramired, -latex] (hidden\xprev2) -- (output2\x);
        
        \node(output22\x)[inner sep=2pt] at (2*\x-1, 3.75) {$t_{\xnext}$};
        \draw[ramired, -latex] (hidden\xprev2) -- (output22\x);
        
    }

    \foreach \x [evaluate={\xprev=int(\x-1)}, evaluate={\xnext=int(\x+1)}]  in {1,...,4}{
        \node(output1\x)[inner sep=2pt] at (2*\x-1, 2.25) {$t_{\x}$};
        \draw[ramired, -latex] (hidden\xprev1) -- (output1\x);
        
        \node(output12\x)[inner sep=2pt] at (2*\x-1, 1.75) {$t_{\xnext}$};
        \draw[ramired, -latex] (hidden\xprev1) -- (output12\x);
        
    }

  \foreach \y in {1, ..., 2}{
    \draw[name=layer\y, draw=none, fill=Apricot, opacity=0.25, rounded corners]  (-0.25, 2*\y-0.33) -- (-0.25, 2*\y+0.33) --  (6.25, 2*\y+0.33) -- (6.25, 2*\y-0.33);
    }
    
  \node(transformerlabel)[draw=none, align=left, rotate=90] at (-1,3) {Transformer Layers};

  \end{tikzpicture}
  \caption{Our example network after adding two predictions per position.}
  \label{figure:multiple_targets}
\end{figure}

\subsection{Positional Embeddings}
\label{section:positional}
In the basic transformer network described in \newcite{NIPS2017_7181}, a sinusoidal timing signal is added to the input sequence prior to the first transformer layer.
However, as our network is deeper (64 layers), we hypothesize that the timing information may get lost during the propagation through the layers.
To address this, we replace the timing signal with a learned per-layer positional embedding added to the input sequence before each transformer layer.
Specifically, the model learns a unique 512-dimensional embedding vector for each of the $L$ context positions within each of $N$ layers, giving a total of $L \times N \times 512$ additional parameters.
We are able to safely use positional embeddings for our task, as we don't require the model to generalize to longer contexts than those seen during training.

\section{Experimental Setup}
\subsection{Datasets}

For evaluation we focus mainly on \texteight{}~\cite{mahoney}.
This dataset consists of English Wikipedia articles, with superfluous content removed (tables, links to foreign language versions, citations, footnotes, markup, punctuation).
The remaining text is processed to use a minimal character vocabulary of 27 unique characters---lowercase letters \texttt{a} through \texttt{z}, and space.
Digits are replaced by their spelled-out equivalents, so ``\texttt{20}'' becomes ``\texttt{two zero}''.
Character sequences not in the range [a-zA-Z] are converted to a single space.
Finally, the text is lowercased.
The size of the corpus is 100M characters.
Following \newcite{mikolov2012subword} and \newcite{zhang2016architectural}, we split the data into 90M characters for train, 5M characters for dev, and 5M characters for test.

To aid in comparison with other recent approaches, we also evaluate our model on \enwikeight{}~\cite{mahoney} which is 100M bytes of unprocessed Wikipedia text, including markup and non-Latin characters.
There are 205 unique bytes in the dataset.
Following \newcite{chung2015gated}, and as in \texteight{}, we split the data into 90M, 5M and 5M for training, dev and test respectively.

\subsection{Training}

\comment{
\todo{Any discussion on less heads working better for some reason?}
}

Compared to most models based on transformers~\cite{NIPS2017_7181,DBLP:journals/corr/abs-1803-05573},
our model is very deep, with 64 transformer layers and each layer using two attention heads.
Each transformer layer has a hidden size of 512 and a filter size of 2048.
We feed our model sequences of length 512.
Each item in the sequence represents a single byte (or equivalently, one character in \texteight{}) which gets replaced by its embedding, a vector of size 512.
We add to the byte embeddings a separate learned positional embedding for each of the 512 token positions, as described in the Positional Embeddings section above.
We do the same addition at each layer activation throughout the network.
The positional embeddings are not shared across the layers.
With two predictions per position, each layer learns to predict 1024 characters.
Because we are primarily interested in predicting the immediately following character (one step away), we halve the loss of predicting characters two steps away.
The prediction layers are logistic regression layers over the full 256 outputs (the number of unique bytes). To demonstrate the generality of the model, we always train and predict over all 256 labels, even on datasets that cover a smaller vocabulary.
Despite this, we found that in practice the model never predicted a byte value outside of the ones observed in the training dataset.

The model has approximately 235 million parameters, which is larger than the number of characters in the \texteight{} training corpus.
To regularize the model, we apply dropout in the attention and ReLU layers with a probability of 0.55.
We use the momentum optimizer with 0.99 momentum.
The learning rate is fixed during training to 0.003.
We train our model for 4 million steps, with each step processing a batch of 16 randomly selected sequences.
We drop the intermediate layer losses consecutively, as described in the Intermediate Layer Losses section above.
Starting from the first layer, after every $62.5$K ($=4$M$\times \frac{1}{2*64}$) steps, we drop the losses introduced by the next layer.
According to this schedule, after training is halfway complete, only the final layer losses are present.

\begin{table}[tb]
    \centering
    \resizebox{0.5\textwidth}{!}{    
    \begin{tabular}{l|rr|r}
        &    \multicolumn{2}{c|}{Parameters ($\times 10^6$)} &   \\ 
        Model & train & inference & bpc \\ \hline
        LSTM~\cite{DBLP:journals/corr/CooijmansBLC16} & - & - & 1.43 \\
        BN-LSTM~\cite{DBLP:journals/corr/CooijmansBLC16} & - & - & 1.36 \\ 
        HM-LSTM~\cite{chung2016hierarchical} & 35 & 35 & 1.29 \\	
        Recurrent Highway~\cite{DBLP:journals/corr/ZillySKS16} & 45 & 45 & 1.27 \\
        mLSTM~\cite{krause2016multiplicative} & 45 & 45 & 1.27 \\
        \smallmodel\ (ours) & 44 & 41 & \textbf{1.18} \\
        \bigmodel~(ours) & 235 & 219 & \textbf{1.13} \\ 
        \midrule
        mLSTM + dynamic eval~\cite{krause2017dynamic} & 45 & - & 1.19 \\        
    \end{tabular}}
    \caption{Comparison of various models on \texteight{} test.}
    \label{tab:final_eval_text8}
\end{table}

\begin{table}[tb]
    \centering
    \begin{tabular}{r|rr|rr}
                & \multicolumn{2}{c|}{bpc} & \multicolumn{2}{c}{Accuracy (\%)}\\ 
        Context & dev & test & dev & test\\ \hline
        32 &  1.25 & 1.34 & 72.8 & 71.1\\
        64 &  1.17 & 1.26 & 74.8 & 73.0\\
        128 & 1.12 & 1.20 & 76.1 & 74.4 \\
        256 & 1.09 & 1.16 & 76.9 & 75.3 \\
        512 & 1.06 & 1.13 & 77.3 & 75.9 \\
    \end{tabular}
    \caption{Bits per character (bpc) and accuracy of our best model on \texteight{} dev and test, for different context lengths.}
    \label{tab:best}
\end{table}

\begin{table}[tb]
    \centering
    \resizebox{0.5\textwidth}{!}{ 
    \begin{tabular}{l|rr|r}
        &    \multicolumn{2}{c|}{Parameters ($\times 10^6$)} &   \\ 
        Model & train & inference & bpb \\ \hline
        FS-LSTM-4~\cite{mujika2017fast} & 47 & - & 1.25 \\ %
        mLSTM~\cite{krause2016multiplicative} & 46 & - & 1.24 \\
        cmix v13~\cite{cmix} & - & - & 1.23 \\ %
        \smallmodel\ (ours) & 44 & 41 & \textbf{1.11} \\ %
        \bigmodel~(ours) & 235 & 219 & \textbf{1.06} \\  %
        \midrule
        mLSTM + dynamic eval~\cite{krause2017dynamic} & 46 & - & 1.08 \\
        
    \end{tabular}
    }
    \caption{Comparison of various models on \enwikeight{} test.}
    \label{tab:final_eval_enwik8}
\end{table}

\subsection{Evaluation}
At inference time, we use the model's prediction at the final position of the final layer to compute the probability of a character given a context of 512 characters.
There is no state passed between predictions as would be the case with RNN models, so for each character predicted we have to process the context from scratch.
Because there is no reused computation from previous steps, our model requires expensive computational resources for evaluation and inference.
We measure the performance of training checkpoints (roughly every 10,000 steps) by evaluating bits per character (bpc) over the entire the validation set, and save the parameters that perform the best.
Our best model is achieved after around 2.5 million steps of training, which takes 175 hours on a single Google Cloud TPU v2. %

\section{Results}

We report the performance of our best model (\bigmodel) on the validation and test sets.
Table~\ref{tab:final_eval_text8} compares our models against several recent results.
On the test set, we achieve a new state of the art, 1.13 bpc.
This model is 5x larger than previous models, which necessitated aggressive dropout rates of 0.55.
For better comparison with smaller models, we also train a smaller model (\smallmodel) with 41M parameters.
This model consists of 12 layers, and trained for 8M steps, with a reduced dropout rate of 0.2.
All other settings were left the same as \bigmodel{}.
Our smaller model still outperforms previous models, achieving 1.18 bpc on the test dataset. Increasing the depth of the network from 12 layers to 64 improved the results significantly, with the auxiliary losses enabling the training to better utilize the depth of the network. Note, our models do not use dynamic evaluation~\cite{krause2017dynamic}, a technique that adjusts model weights at test time by training on test data.

Table \ref{tab:best} shows the performance of our model given different context sizes.
We are able to achieve state-of-the-art results once the context increases beyond 128 characters, with the best performance of 1.06 bpc at 512 characters.
As expected, the model performs better when it is given more context. However this trend levels off after 512 characters; we do not see better results using a context of 1024.

Using the same hyperparameters and training procedure for \texteight{}, we also train and evaluate the \smallmodel{} and \bigmodel{} architectures on \enwikeight{} (see Table \ref{tab:final_eval_enwik8}).
Note, several previous authors discuss ``bits per character'' on \enwikeight{} but are in fact reporting bits per byte.
Without retuning for this dataset, our models still achieve state-of-the-art performance.

\subsection{Ablation Experiments}
\label{section:ablation}

To better understand the relative importance of the several modifications we proposed, we run an ablation analysis.
We start from our best model \bigmodel{} and then remove one modification at a time.
For example, when we disable \emph{Multiple Positions}, the model gets trained with only the last position loss for each layer.
This corresponds to calculating \{$L(t_4\mid t_{0:3})$, $L(t_5\mid t_{0:3})$\} in the example shown in Figure \ref{figure:multiple_targets} for both the first and the second layers.
When disabling \emph{Positional Embeddings}, we add the default transformer sinusoidal timing signal before the first layer.

\begin{table}[ht]
    \centering
    \begin{tabular}{l|c|c}
        Description & bpc & $\Delta$bpc \\ \hline
        \bigmodel{} (Baseline) & 1.062 & - \\
        \bigmodel{} w/out Multiple Positions & 2.482 & 1.420 \\
        \bigmodel{} w/out Intermediate Layer Losses & 1.158 & 0.096 \\
        \bigmodel{} w/out Positional Embeddings & 1.069 & 0.007 \\
        \bigmodel{} w/out Multiple Targets & 1.068 & 0.006 \\
        
        \bigmodel{} w/ SGD Optimizer & 1.065 & 0.003 
    \end{tabular}
    \caption{Evaluation of \bigmodel\ on \texteight\ dev with context set to 512. Disabling each feature or loss lowers the quality of the model. The biggest win comes from adding multiple positions and intermediate layers losses.}
    \label{tab:my_label}
\end{table}

\comment{
\begin{table}[ht]
    \centering
    \begin{tabular}{l|r}
        Description & bpc \\ \hline
        \bigmodel{} &  1.062  \\
        \midrule
        \bigmodel{} without Multiple Positions & 2.482 \\
        \bigmodel{} without Intermediate Layer Losses & 1.158 \\
        \bigmodel{} without Multiple Targets & 1.068 \\
        \bigmodel{} without Positional Embeddings & 1.069 \\
        \bigmodel{} with SGD Optimizer & 1.065
    \end{tabular}
    \caption{Evaluation of \bigmodel\ on \texteight\ dev with context set to 512. Disabling each feature or loss lowers the quality of the model. The biggest win comes from adding multiple positions and intermediate layers losses.}
    \label{tab:my_label}
\end{table}
}

For the ablation experiments, we reuse the hyperparameters from our best model to avoid a prohibitively expensive parameter search for each ablation.
The only exception is the SGD experiment, where we vary the learning rate.
The analysis shows that the biggest advantage comes from multiple positions and intermediate layers losses.
Predicting all the intermediate positions leads to significant speed up in convergence, since the model sees more effective training examples per batch.
Adding losses at the intermediate layers acts in the same spirit by forcing more predictions per training step.

Finally, we replace momentum with SGD as our optimizer, using a range of learning rates (0.3, 0.1, 0.03, 0.01, 0.003, 0.001).
This ablation shows that SGD produces competitive models, with learning rate 0.1 giving the best performance.
Despite the depth of our network, SGD is able to train the network efficiently with the help of our auxiliary losses.

\subsection{Comparison with Word-Level Models}
To understand how byte-level language models perform in comparison to word-level language models, we train \bigmodel{} on the \lmoneb{} corpus \cite{chelba2013one}.
For \lmoneb, we use the standard train/test split of the preprocessed corpus, where out-of-vocab words have been replaced with \texttt{UNK}, to allow comparison to previous work on word and word-piece models.
We report word perplexity (\texttt{ppl}) by converting bits-per-byte (\texttt{bpb}) into \texttt{ppl}\footnote{For this test set, $ppl = 2 ^ {bpb * 826189 / 159658}$, where 826,189 is the number of bytes and 159,658 is the number of tokens.}.
During training we use the second shard (01) of the heldout dataset as a dev set, as the first shard (00) is the test.
Given this is a significantly larger dataset than \texteight{}, we set all dropouts to zero.
Table \ref{tab:lm1b} shows a gap in performance between the two classes of language models.
This comparison can serve as a starting point for researching possible ways to bridge the gap.

\begin{table}[tb]
    \centering
    \begin{tabular}{ll|rr}
        Type & Model & bpb & ppl \\ \hline
        Word &    \newcite{DBLP:journals/corr/JozefowiczVSSW16} & - & 23.7 \\
        Byte & \bigmodel{} & 1.03 & 40.6 \\
    \end{tabular}
    \caption{Performance of \bigmodel{} on the \lmoneb{} test set.}
    \label{tab:lm1b}
\end{table}

\subsection{Qualitative Analysis}

To probe the strengths and weaknesses of our best model (\bigmodel), we run the model forward, starting with the seed sequence of 512 characters in Figure \ref{fig:completions}, taken from the \texteight{} test set.
Figure \ref{fig:predictions} shows several per-character metrics for the model's predictions over the true continuation of this seed text.
At each position, we measure i) the model's prediction entropy in bits across all 256 output classes, ii) its loss---the negative log probability of the target label, i.e.~the ``bits per character'' for this position, and iii) the rank of the target in the list of output classes sorted by likelihood.
Unsurprisingly, the model is least certain when predicting the first character of a word, and becomes progressively more confident and correct as subsequent characters are seen.

\begin{figure}[tb]
\hspace{-15pt}
\begin{tabular}{p{0.5\textwidth}}
    \multicolumn{1}{c}{\textbf{\small Seed}} \\ \vspace{-8.0pt}
    \corpus{mary was not permitted to see them or to speak in her own defence at the tribunal she refused to offer a written defence unless elizabeth would guarantee a verdict of not guilty which elizabeth would not do although the casket letters were accepted by the inquiry as genuine after a study of the handwriting and of the information contained therein and were generally held to be certain proof of guilt if authentic the inquiry reached the conclusion that nothing was proven from the start this could have been pr} \\
    \multicolumn{1}{c}{\textbf{\small Word Completions}} \\ \vspace{-8.0pt}
    \corpus{proven, proved, proof, prevented, presented, problematic, probably, provided, practical, provoked, preceded, predicted, previously, presumed, praised, proposed, practicable, produced, present, preserved, precisely, prior, protected, probable, prompted, proofed, properly, practiced, prohibited, profound, preferable, proceeded, precise, predictable, practically, prevalent}
\end{tabular}
 \caption{A seed sequence of 512 characters taken from the \texteight{} test set, and all word completions assigned cumulative probability above 0.001 to follow the seed, in order from most likely (0.529) to least likely (0.001).}
 \label{fig:completions}
\end{figure}

\begin{figure*}[ht]
    \centering
    \includegraphics[width=\textwidth]{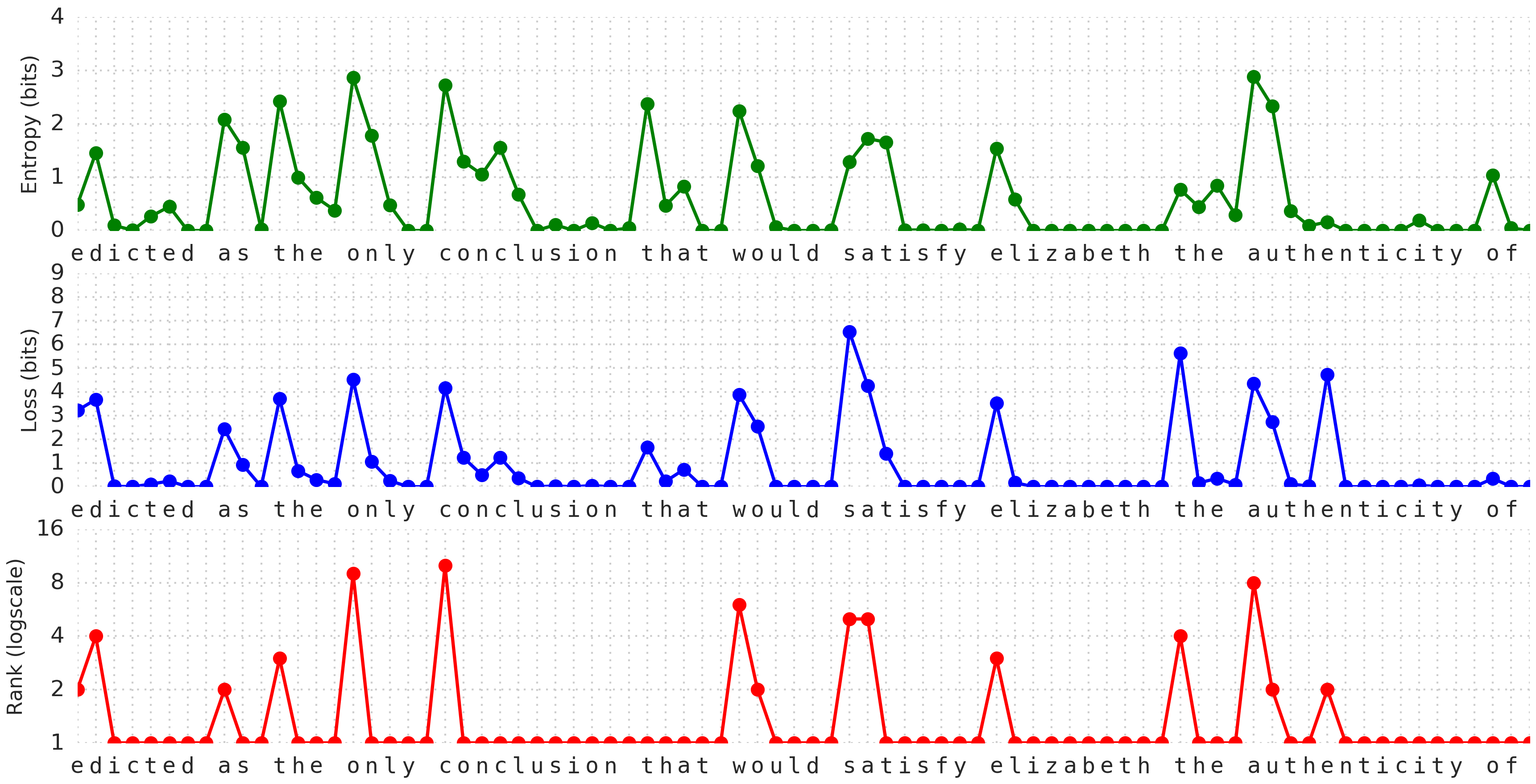}
    \caption{Per-character entropy, loss and rank assigned by \bigmodel{} after seeding on the 512 character sequence from Figure \ref{fig:completions}.}
    \label{fig:predictions}
\end{figure*}

To investigate the degree to which our model prefers actual English words over non-existent words, we compute the likelihood the model assigns to all continuations after the seed. We cut off continuations when they reach a space character, or when the total probability of the continuation falls below 0.001. Figure \ref{fig:completions} shows the entire set of word completions, in order of probability, where the initial \texttt{pr-} from the seed is repeated for readability. Note that these are all real or plausible (\texttt{proofed}) English words, and that even short but bad continuations like \texttt{prz} are assigned a lower cumulative probability than long realistic word completions like \texttt{predictable}.

We expect that the transformer self-attention should make it easy for our model to copy sequences observed in the context over long distances (up to the context size of 512 characters). To test this expectation, we corrupt the seed and continuation from above by introducing a fake name \texttt{zjakdmu bmijwxn}. Specifically, we change the first occurrence of \texttt{elizabeth} in the seed to \texttt{zjakdmu bmijwxn}, and the second occurrence to \texttt{she}. Similarly, in the continuation, we change \texttt{elizabeth} to \texttt{zjakdmu bmijwxn}. The resulting distance between the two occurrences of the fake name is 434 characters.

\begin{figure*}[ht]
    \centering
    
    \begin{subfigure}[b]{\textwidth}
        \centering
        \includegraphics[width=\textwidth]{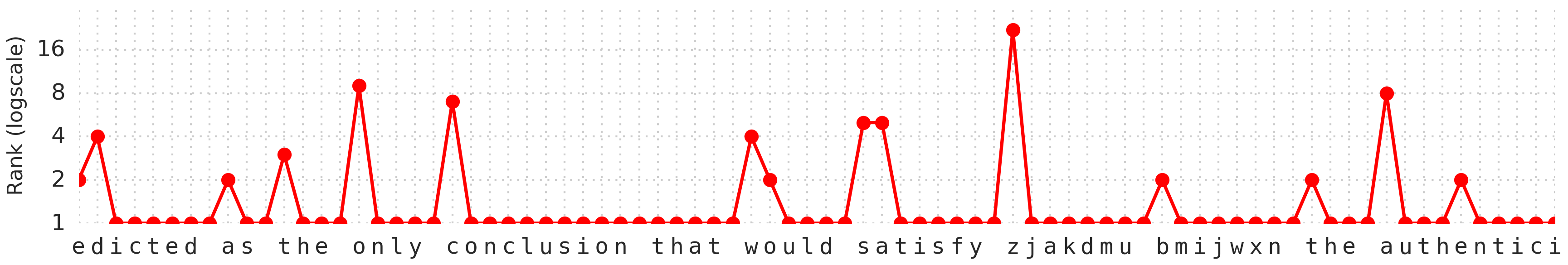}
        \caption{Continuing after the modified seed (including the fake name 434 characters away).}
        \label{fig:copy1}
    \end{subfigure}

    \begin{subfigure}[b]{\textwidth}
        \includegraphics[width=\textwidth]{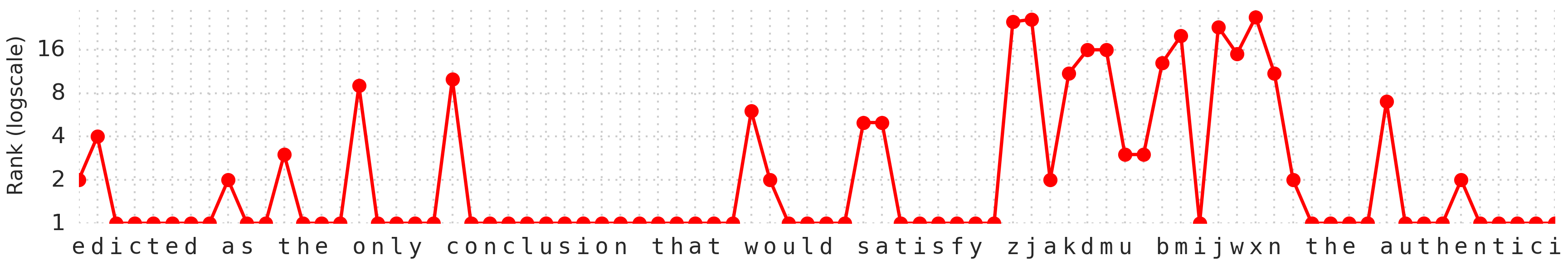}
        \caption{Continuing after the original seed from Figure \ref{fig:completions}.}
        \label{fig:copy2}
    \end{subfigure}
    \caption{Per-character rank assigned by \bigmodel{} to a fake continuation, after being seeded on either (a) the fake context where \texttt{elizabeth} is replaced with \texttt{zjakdmu bmijwxn}, or (b) the original context.}
\end{figure*}

Figure \ref{fig:copy1} confirms that the model can successfully copy over this long distance. While the initial \texttt{z} in \texttt{zjakdmu} is unexpected, the model immediately chooses to copy the remainder of this word from the context, as opposed to predicting any real \texttt{z-} words learned during training. Similarly, while the model is somewhat unsure whether the fake surname \texttt{bmijwxn} will appear (assigning the initial \texttt{b} a rank of two), it immediately picks up on the correspondence after the \texttt{b} is observed, correctly predicting the remainder of the fake surname.

For comparison, Figure \ref{fig:copy2} shows how the model would rank the targets in our fake continuation if the original seed with \texttt{elizabeth} were used. This confirms that the fake name is not predictable based on knowledge gained through training, and is indeed being copied from the preceding context.

\comment{
As further corroboration, we observe that the learned attention masks within the transformer self-attention layer often display ``diagonal streaks'', as in Figure \ref{fig:copy_attention}, where a later sequence attends backward one-to-one onto an earlier sequence.

\begin{figure}[!htbp]
    \centering
    \includegraphics[scale=0.6]{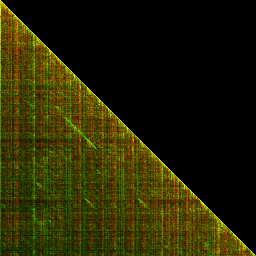}
    \caption{The multihead attention mask within a transformer layer of an early development model, showing which query positions (on the y-axis) are attending to which key positions (on the x-axis), for a single example from the validation set. The black triangle reflects the restriction that each position can only attend leftward. The diagonal green streaks indicate that one attention head has learned to attend backward in an aligned one-to-one mapping onto a previous sequence. Across several such examples, we observe that each position is attending back to the character \emph{following} itself in a previous occurrence of the same phrase.}
    \label{fig:copy_attention}
\end{figure}
}

\begin{table}[h!]
\begin{tabular}{p{0.45\textwidth}}
\multicolumn{1}{c}{\textbf{\small Seed}} \\\hline \vspace{-8.0pt}
\corpus{\tvtick{}Computational neuroscience\tvtick{}\ is an interdisciplinary field which draws on [[neuroscience]], [[computer science]], and [[applied mathematics]].  It most often uses mathematical and computational techniques such as computer [[simulation]]s and [[mathematical model]]s to understand the function of the [[nervous system]].\newline{}The field of computational neuroscience began with the work of [[Andrew Huxley]], [[Alan Hodgkin]], and [[David Marr]].  The results of Hodgkin and Huxley\svtick{}s pioneering work in developing } \\\hline
\end{tabular}
\begin{tabular}{l|p{0.35\textwidth}}
Sample 1 & {\small \emph{computational neuroscience were chronicled in \dvtick[[Is Mathematics Anything I Could Learn?]]\dvtick. (ISBN 0826412246).  Computational }} \\
Sample 2 & {\small \emph{neuroscience concerned neurological auraria and the inherited ability to communicate and respond to environmental destruction - }} \\
Sample 3 & {\small \emph{the model were published in 1982 and 1983 respectively, and the subsequent work on the field began its graduate program with [[M}} \\ \vspace{-6pt}
& \\
Truth & {\small \emph{the voltage clamp allowed them to develop the first mathematical model of the [[action potential]]. David Marr\svtick{}s work focuses on}}
\end{tabular}
\caption{Samples generated by \bigmodel{}, seeded with text from the \enwiknine{} dev set, using a sampling temperature of 1.0.}
\label{tab:gen}
\end{table}

\subsection{Generation}
For generating samples using our language model, we train on a larger and less processed dataset, \enwiknine{} \cite{mahoney}.
We split \enwiknine{} into 900M, 50M and 50M for training, dev and test.
Using the dev dataset to tune our dropout, we find that dropout=0.1 performs the best.
On the test dataset, \bigmodel{} achieves 0.85 \texttt{bpb}.
Table \ref{tab:gen} shows different generated samples following the seed text, using a sampling temperature of 1.0.

\section{Related Work}
Character-level modeling has shown promise in many areas such as sentiment analysis~\cite{DBLP:journals/corr/RadfordJS17}, question answering~\cite{kenter2018byte} and classification~\cite{DBLP:journals/corr/ZhangZL15}, and is an exciting area due to its simplicity and the ability to easily adapt to other languages.
Neural network based language modeling has been heavily researched since its effectiveness was shown by \newcite{Bengio:2003:NPL:944919.944966}.
By far, the most popular architecture in this area is the RNN and variants, first studied in \newcite{conf/interspeech/MikolovKBCK10}.

Much of the progress in this area has been made by mitigating the vanishing gradients problem~\cite{hochreiter2001gradient} by architectures such as LSTMs~\cite{LSTM}, GRU~\cite{DBLP:journals/corr/ChoMGBSB14}, Recurrent Highway Networks~\cite{DBLP:journals/corr/ZillySKS16}, Unitary RNNs~\cite{DBLP:journals/corr/ArjovskySB15} and others. This is an issue that transformers do not have, due to attention allowing short paths to all inputs.
Methods of normalizing activation functions, such as Batch Normalization~\cite{DBLP:journals/corr/IoffeS15,DBLP:journals/corr/abs-1708-02182} and Layer Normalization~\cite{2016arXiv160706450L} have also demonstrated improvements on language modeling tasks. As with this work, progress has been made with discovering ways to regularize sequential architectures, with techniques such as Recurrent Dropout~\cite{DBLP:journals/corr/ZarembaSV14,2015arXiv151205287G} and Zoneout~\cite{krueger2016zoneout,rocki2016surprisal}.

A closely related architecture is the Neural Cache Model~\cite{DBLP:journals/corr/GraveJU16}, where the RNN is allowed to attend to all of its previous hidden states at each step. Another similar model is used in~\cite{DBLP:journals/corr/DanilukRWR17} where a key-value attention mechanism similar to transformers is used. Both approaches show improvements on word level language modeling. Memory Networks~\cite{DBLP:journals/corr/WestonCB14} have a similarity to the transformer model in design as it also has layers of attention for processing a fix memory representing the input document and has been shown to be effective for language modeling in \cite{sukhbaatar2015end}. ByteNet~\cite{kalchbrenner2016neural}, which is related but uses layers of dilated convolutions rather than attention, showed promising results on byte level language modeling. Gated Convolutional Networks~\cite{DBLP:journals/corr/DauphinFAG16} was an early non-recurrent model to show superior performance on word level language modeling.

Language models are not usually very deep due to computational constraints of training RNNs, and this also limits the number of parameters. The transformer architecture allowed us to build very deep (64 layer) models with a large number of parameters. A recent CNN model for text classification~\cite{DBLP:journals/corr/ConneauSBL16} at 29 layers is considered deep in the NLP community. A Sparsely-Gated Mixture-of-Experts Layer~\cite{DBLP:journals/corr/ShazeerMMDLHD17} allowed language modeling experiments with a greatly increased number of parameters by only accessing a small portion of parameters every time step, showing a reduction in bits per word. In Exploring the Limits of Language Modeling~\cite{DBLP:journals/corr/JozefowiczVSSW16}, an increase in the number of parameters was achieved by mixing character-level and word level models, using specialized softmaxes and using a large amount of computational resources to train. IndRNN~\cite{DBLP:journals/corr/abs-1803-04831} uses a simplified RNN architecture that allows deeper stacking with 21-layers, achieving near SOTA character-level language modeling. Fast-Slow Recurrent Neural Networks~\cite{mujika2017fast} also achieved near SOTA by increasing the number of recurrent steps for each character processed.

\section{Conclusion}
Character language modeling has been dominated by recurrent network approaches.
In this paper, we show that a network of 12 stacked transformer layers achieves state-of-the-art results on this task.
We gain further improvements in quality by deepening the network to 64 layers, utilizing capacity and depth efficiently.
The use of auxiliary losses at intermediate layers and positions is critical for reaching this performance, and these losses allow us to train much deeper transformer networks.
Finally, we analyze the behavior of our network and find that it is able to exploit dependencies in structure and content over long distances, over 400 characters apart.

\fontsize{9.5pt}{10.5pt} \selectfont 
\bibliographystyle{aaai}
\bibliography{references}
\end{document}